\pdfoutput=1
\documentclass[10pt]{article}
\usepackage[letterpaper]{geometry}
\usepackage{hicss}
\usepackage{times}
\usepackage[none]{hyphenat}
\usepackage{url}
\usepackage{latexsym}
\usepackage{indentfirst}
\usepackage{graphicx}
\graphicspath{{images/}}
\usepackage[
backend=biber,
   style=apa,
 ]{biblatex}
\usepackage{amsmath,amssymb,amsfonts}
\usepackage[linesnumbered,ruled,lined]{algorithm2e}
\SetKwInput{KwInput}{Input}                
\SetKwInput{KwOutput}{Output}
\SetKwData{KwDat}{Variables}
\usepackage{caption}
\usepackage{subcaption}
\usepackage[export]{adjustbox}
\usepackage{inconsolata}
\usepackage[most]{tcolorbox}
\usepackage{booktabs}
\usepackage{multirow}

\usepackage[noframe]{showframe}
\usepackage{framed}
\usepackage{lipsum}
\renewenvironment{shaded}{%
  \MakeFramed{\advance\hsize-\width \FrameRestore\FrameRestore}}%
 {\endMakeFramed}
\definecolor{shadecolor}{gray}{0.75}

\title{Narrating Causal Graphs with Large Language Models}

\author{Atharva Phatak \\
  Lakehead University, Thunder Bay, ON, Canada \\
  {\underline{phataka@lakeheadu.ca}}
	\\ \\
	Ameeta Agrawal, Aravind Inbasekaran\\
	Portland State University, Portland, OR, USA \\
	{\underline{\{ameeta,ai23\}@pdx.edu}}
	\\\And 
	Vijay K. Mago\\
	York University, Toronto, ON, Canada\\
	{\underline{vmago@yorku.ca}}
	\\ \\
	Philippe J. Giabbanelli\\
	Miami University, Oxford, OH, USA\\
	{\underline{giabbapj@miamioh.edu}} }

\date{}

\begin{document}
\maketitle
\begin{abstract}
The use of generative AI to create text descriptions from graphs has mostly focused on knowledge graphs, which connect concepts using facts. In this work we explore the capability of large pretrained language models to generate text from causal graphs, where salient concepts are represented as nodes and causality is represented via directed, typed edges. The causal reasoning encoded in these graphs can support applications as diverse as healthcare or marketing. Using two publicly available causal graph datasets, we empirically investigate the performance of four GPT-3 models under various settings. Our results indicate that while causal text descriptions improve with training data, compared to fact-based graphs, they are harder to generate under zero-shot settings. Results further suggest that users of generative AI can deploy future applications faster since similar performances are obtained when training a model with only a few examples as compared to fine-tuning via a large curated dataset.
\end{abstract}

\subsubsection*{Keywords: Causal Map, Generative AI, GPT, Pre-Trained Large-Scale Language Model} 

\section{Introduction}
\label{sec:intro}
Large-scale pre-trained language models (LLMs) such as ChatGPT have recently been at the forefront of generative AI. By accomplishing a variety of tasks, these models save time for human users. They provide an accessible technology, as users do not require expertise in natural language processing (NLP). For example, GPT-based solutions can be integrated in information systems to create summaries (\cite{ma2023impressiongpt}), which is faster than asking humans to read a large textual input and does not require expertise in abstractive summarization algorithms. There is also a strong interest in using these models to perform \textit{causal reasoning}, as it has potential to improve both customer experience and intention to use chatbots in areas such as healthcare information systems (\cite{yu2021dynamic}) or marketing (\cite{bialkova2023want}). In a classic example, if a user asks ``What will happen to my headache if I take an aspirin?'' then the chatbot needs a causal model to suggest that the headache will be gone (\cite{bishop2021artificial}). However, AI practitioners have noted that causality in LLMs is a nascent research field, so companies may currently struggle to effectively integrate LLMs by treating them over-confidently as human-level intelligence (\cite{zhang2023causality}). This has important implications, as exemplified by a recent case in which a user found the causal reasoning of a chatbot so convincing that he followed it to the letter, even when the chatbot encouraged him to commit suicide (\cite{Biztech}). It is thus essential to assess and improve causal reasoning in LLMs (\cite{kiciman2023causal}), such that we evaluate the limitations of their application and potentially add causal information to their training set.

\begin{figure}
\centering
\includegraphics[width=0.8\linewidth]{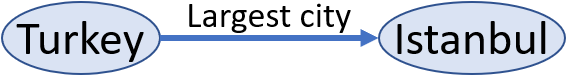}
  \caption{Sample graph from WebNLG dataset.}
	\label{fig:web_nlg}
	\end{figure}

Given that causality focuses on connecting antecedents to their consequences, a directed graph is a frequently used structure in causal research. Commonly employed types of graphs for generative AI include the following: \textit{ontologies} (which possess attributes, classes, and events), where edges can be labeled as a subclass or a cause; 
\textit{knowledge graphs} or semantic networks as used in WebNLG (\cite{wang2023improving}), where edges are labeled by words (see Figure~\ref{fig:web_nlg}); and \textit{causal maps} (\cite{Anish2022}), where edges are typed/labeled as positive or negative (see Figure~\ref{fig:Obesity_causal}). In this paper, we assess how much (if any) data is necessary to a LLM in generating sentences with the right type of causal effect. That is, we use GPT-3 to transform causal maps into textual outputs that must have the appropriate causal increase or causal decrease. 

Generating sentences from a graph is known as \textit{graph-to-text} generation, which is a subtask of data-to-text generation. Recent studies in graph-to-text have shown that a causal graph could be transformed into fluent textual outputs (\cite{Anish2022}), as captured by both automatic metrics and manual assessments (both of which are also employed in the present study). However, these works also relied on many additional pairs of examples (i.e., input graph and desired text) in order to train GPT-3 both on the application domain and on the causality encoded in the graphs. By examining whether this work-intensive fine-tuning can be reduced (few-shot learning) or eliminated altogether (zero-shot), our work contributes to lessening the burden on users and thus makes it possible to turn the wide array of available causal maps (\cite{wang2023identifying}) into text.



The main contributions of our work are twofold:
\begin{itemize}
    \item We evaluate the possibility of transforming causal graphs to text \textit{without} having to specify causality. Our results are provided on two causal datasets, three different training settings (zero-shot, few-shot, and fine-tuned), and four GPT-3 models.
    \item We contrast results using both automatic performance metrics and human evaluations.
\end{itemize}

The remainder of this paper is organized as follows. In section~\ref{sec:related}, we succinctly review the context leading to the creation of causal maps and provide a brief background on LLMs. We present our methods and datasets in section~\ref{sec:methods}. Our results are presented in section~\ref{sec:results}, including our performances and an assessment of the differences between our approach and prior works. These results are discussed in section~\ref{sec:discussion} to address our central question: can a large language model such as GPT-3 act as a causal learner, or do we need to manually convey causation? 

\section{Background}\label{sec:related}

\subsection{Causal Maps}

A \textit{causal map} is a representation of a system as a graphical model (\cite{de2017mapping}), where salient concepts are captured as labeled nodes (e.g., `financial insecurity', `homelessness'), and causality is represented via directed, typed edges (e.g., financial insecurity $\xrightarrow{\text{+}}$ homelessness). While a \textit{knowledge graph} describes factual knowledge in the form of relations between entities, a causal map is a specific case in which relations can take two values and knowledge is \textit{subjective} since it provides the perspectives of an individual. These maps are often produced in the context of \textit{participatory modeling} (\cite{quimby2022participatory}), where participants (e.g., stakeholders, experts, community members) share their views as individual causal maps which are then aggregated to obtain a comprehensive view. Although the method of causal mapping is often chosen because it allows to elicit perspectives in a transparent manner with participants (\cite{voinov2018tools}), the \textit{product} may no longer be transparent as participants struggle to interact with maps (\cite{giabbanelli2023human}). This has motivated prior works in converting maps into text, as a more universally accessible format (\cite{Anish2022}).




\begin{figure}
\centering
\includegraphics[width=\linewidth]{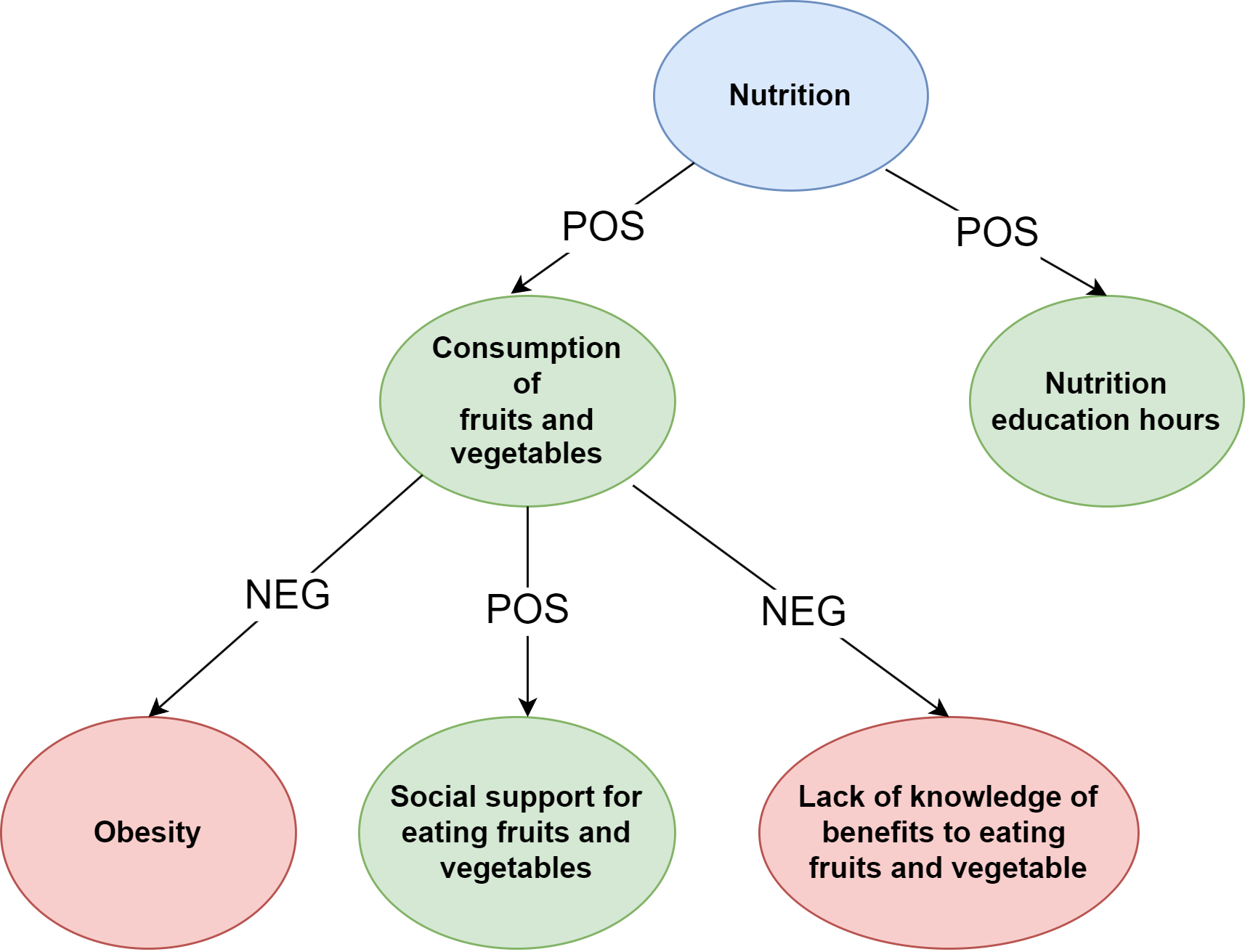}
    \caption{Sample graphs from Obesity dataset. The linearized representations of instances would be: \texttt{$<$S$>$ $<$H$>$ nutrition $<$POS$>$ $<$T$>$ consumption of fruits and vegetables$<$H$>$ nutrition $<$POS$>$ $<$T$>$ nutrition education hours$<$H$>$ consumption of fruits and vegetables $<$NEG$>$ $<$T$>$ obesity$<$H$>$ consumption of fruits and vegetables $<$POS$>$ $<$T$>$ social support for eating fruits and vegetables$<$H$>$ consumption of fruits and vegetables $<$NEG$>$ $<$T$>$ lack of knowledge of benefits to eating fruits and vegetables $<$E$>$}.}
    \label{fig:Obesity_causal}
  \end{figure}
	
\subsection{Graph-to-text Generation}
Early neural models to generate text descriptions from graphs were mostly fully supervised requiring large annotated datasets, and included  architectures such as sequence-to-sequence, graph transformer (\cite{wang2020amr}), heterogeneous graph transformer (\cite{yao2020heterogeneous}), and graph encoder-decoder (\cite{shi2020g2t}). Recent progress on generative pre-trained language models (PLMs) has achieved impressive results in graph-to-text generation. (\cite{mager2020gpt}) were the first to employ a decoder-only PLM (GPT-2) to transform Abstract Meaning Representation graphs (directed trees that form whole sentences) into text (\cite{radford2018improving}). This was followed by (\cite{ribeiro2020investigating}) who investigated the impact of different task-adaptive pretraining strategies for two encoder-decoder PLMs including the popular BART (Bidirectional and Auto-Regressive Transformer) and T5 models. In particular, they showed that approaches based on PLMs outperformed those explicitly encoding graph structure. 

An emerging research area has been to control the generated text such that it expresses a set of user-desired attributes. (\cite{hu2021causal}) focus on controllable text generation from a causal perspective with the primary objective of reducing bias in the text generated by various conditional models. (\cite{li2021guided}) proposed causal generation models utilizing transformers. They constructed a corpus called CausalBank, which consists of 314 million cause-effect pairs. This corpus was used to train the model, enabling the generation of cause and effect relationships given initial words in a sentence. Note that there are now several studies that examine causality in generative AI and obtain seemingly contradictory results. This is partly explained by the \textit{different tasks} used across studies (\cite{kiciman2023causal}). Some studies strive to generate text that learns the whole causal graph (i.e., counterfactuals) while others may provide all pairs of related concepts and only expect the generator to correctly identify which concept is the antecedent and which one is the consequent. We thus emphasize the importance of being specific with respect to the causal task of interest (Section~\ref{sec:description}).


Recent work has also focused on data-to-text generation under few-shot setting (\cite{li2021few}), zero-shot setting (\cite{kasner2022neural}) and any-shot setting (\cite{xiang2022asdot}). These different settings are also explored in the present manuscript. Among notable works, (\cite{chen2020kgpt}) proposed a knowledge-grounded pre-training framework and evaluated it under fully-supervised, zero-shot, and few-shot settings. (\cite{hoyle2020promoting}) explored scaffolding objectives in PLMs (T5) and showed gains in low-resource settings. 

Most of the previous methods have studied graph-to-text generation through widely-used datasets such as WebNLG (\cite{gardent2017webnlg}), hence we include this dataset in our study to allow for comparison. Complementary to prior work, we focus on {\em causal datasets} which contain facts that are often not explicitly stated in knowledge graphs.




\section{Methods}
\label{sec:methods}
\subsection{Problem Description}
\label{sec:description}
A causal graph $G = (V, E)$ consists of a set of entities (the nodes $V$) and relations (the edges $E$). Each entity $v \in V$ has a label, which can be composed of multiple words and is usually a noun (e.g., nutrition, consumption of fruits and vegetables). Each relation is directed since it encodes causality and it can be of only two types (positive, negative). Given a causal graph $G$, our goal is to generate a set of sentences $S=\{s_1, s_2, \ldots, s_n\}$ that describes the graph in natural language text. For example, given the graph in Figure~\ref{fig:Obesity_causal}, sentences could be as follows:
\begin{shaded}
Increasing nutrition education improves the consumption of fruits and vegetables, which prevents obesity and provides social support to consume such foods. As individuals eat more fruits and vegetables, there is also a lesser lack of knowledge about the associated benefits. Another consequence of a rise in nutrition education is that more hours will be spent on this topic.
\end{shaded}

The simplest solution would be to turn every edge $A \xrightarrow{\text{type}} B$ into a sentence \textit{A increases/decreases B}, which would achieve perfect scores in all automatic metrics since the output would be grammatically correct, contain the data present in each edge, and does not create noisy data (i.e., hallucinate). However, such a template-based approach would be unpleasant for humans to read. In contrast, Generative AI solutions are expected to express causality in diverse ways (e.g., improves, lessens, augments), combine edges into single sentences when there is a shared root node, or express a sequence of edges (i.e., a path) in one sentence to form a logical thread. The downside is a potential decrease in various metrics, particularly as approximations may ignore the type of causality or some edges entirely, or hallucinate due to the reliance on deep neural networks.

\subsection{Data Pre-Processing}
Since a text description is linear (read from left-to-right) but a graph can contain \textit{cycles}, the graph is first decomposed into a series of acyclic components. The decomposition is not a simplification, as the union of all components should be equal to the input graph. Consequently, any loss of information in the text output would be attributable to the NLG step rather than to pre-processing. To preserve information while decomposing the graph into acyclic components, it may be necessary to include some nodes or edges in multiple components. For example, consider $A \rightarrow B \rightarrow C \rightarrow A$. This could be split into $A \rightarrow B \rightarrow C$ and $C \rightarrow A$, hence $C$ appears in both components. This redundancy is adequate for NLG tasks, since sentences on a given topic could also repeat some concepts or important causal statements. In our example from section 3.1, nutrition education was present in multiple sentences. Prior research has provided algorithms to obtain such a `linearized representation' of a graph and showed that small graph sizes (less than 10 nodes) perform best for NLG tasks (\cite{Anish2022}). We thus used the algorithm published in this prior work and followed their recommendation to create small components. We emphasize that the focus of our work is on generating sentences without having to specify the causality, and with lesser or no training data.


\begin{figure}
    \centering
    \includegraphics[scale=0.4]{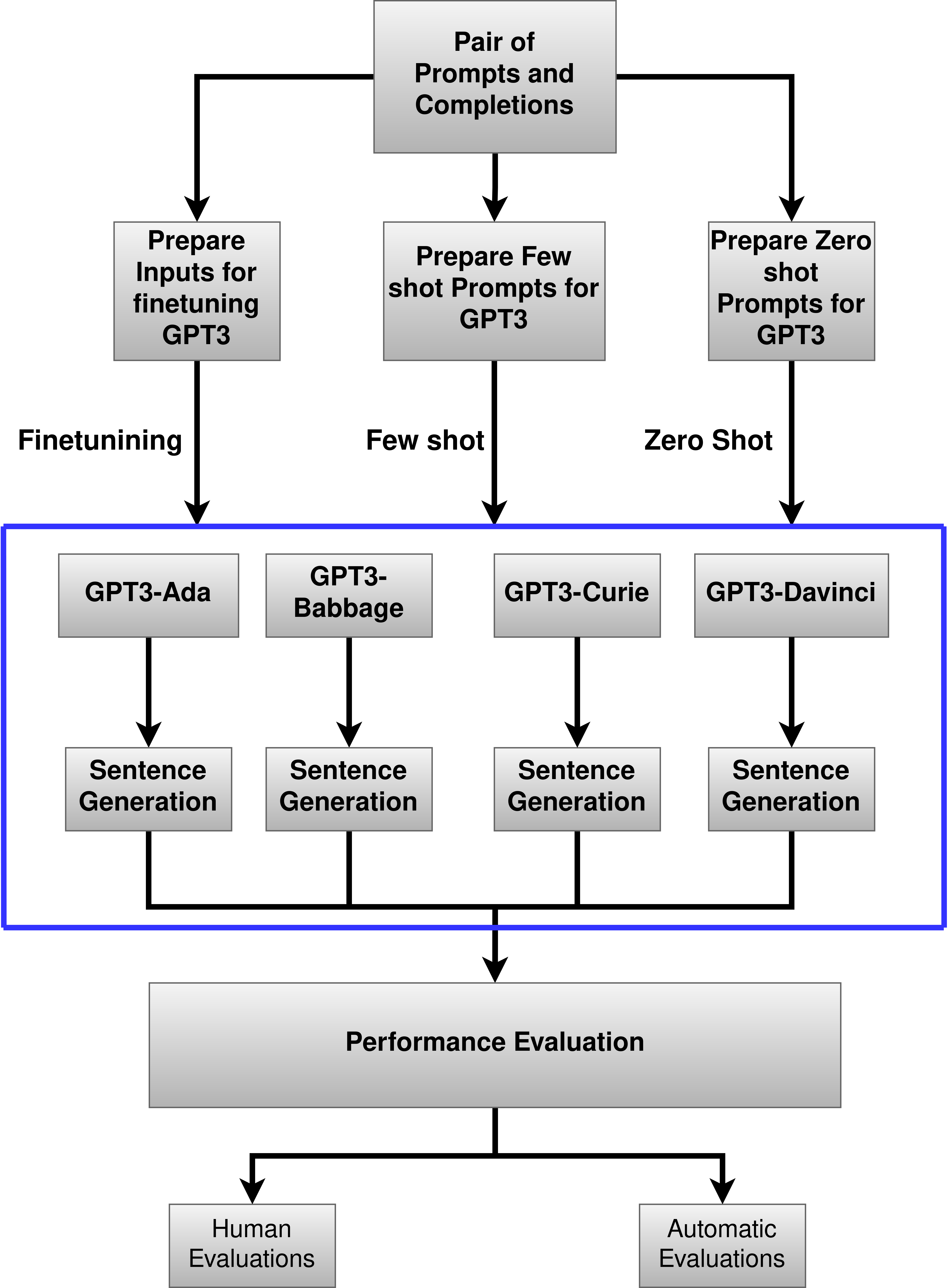}
    \caption{We had three training settings (fine-tuned, few shot, zero shot) and four variants of GPT-3 models.}
    \label{fig:workflow-image}
\end{figure}

\subsection{Experimental Approach}

While many PLMs can be used for text generation, we use four variants of OpenAI's GPT-3 models\footnote{\url{https://openai.com/api/}} (\cite{gpt3-paper}): Davinci (175 billion parameters), Curie (6.7B), Babbage (1.3B), and Ada (350M). These models are used via a $2 \times 3$ experimental approach consisting of 2 versions of the input data (with/without expressed causality) and 3 learning settings (fine-tuning, few shot, zero shot), detailed below. The pseudocode for our methods is provided in the Appendix.

We considered two versions of the input. The first version is produced directly by the pre-processing step explained in section 2.2, which \textbf{includes causal tags}. In our modified second version, we \textbf{exclude causal tags} by replacing $\langle POS \rangle$ and $\langle NEG \rangle$ with a generic causal connector. These two input versions allow us to test whether the generative AI is capable of inferring the type of causality. In each version, we added a pipe character ($|$) as a delimiter between each entity relation, such that the edges were clearly segmented in the input.

We also evaluated three training settings (see Figure~\ref{fig:workflow-image}), to examine how much data was necessary for a generative AI to infer causality. The most common setting is to create a \textbf{fine-tuned} model by training a PLM using a task-specific dataset, which we constructed for this experiment as detailed below. The effect of fine-tuning is that a model improves its performances by updating its weights. The OpenAI API recommends to fine-tune the models for a short amount of epochs. We observed that smaller GPT models (Ada, Babbage) had poor results for 1 or 2 epochs, hence we used 5 epochs for all models to guarantee that results reach a plateau.

The second setting reduces the training set via a \textbf{few-shot} approach. Instead of a large number of training samples, only $k$ samples are chosen, where $k$ depends on the model's context size. That is, GPT has a maximum input length limit (i.e., context length of 2048 tokens), so the set of all examples must fit within this limit; the more context is required when providing an example, the fewer examples can fit. Previous works have used very different amounts of examples depending on their length. For instance, one study used 5 examples (\cite{agrawal2022large}), another varied from 1 to 16 examples (\cite{yang2022empirical}), and a third tested up to 100 examples before running out of tokens (\cite{moradi2021gpt}). In our case, we use 3 examples randomly selected from the wider pool used in fine-tuning, hence we employ a \textit{strict} few-shot setting. 




\begin{figure}[!ht]
    \centering
    \includegraphics[width=\linewidth]{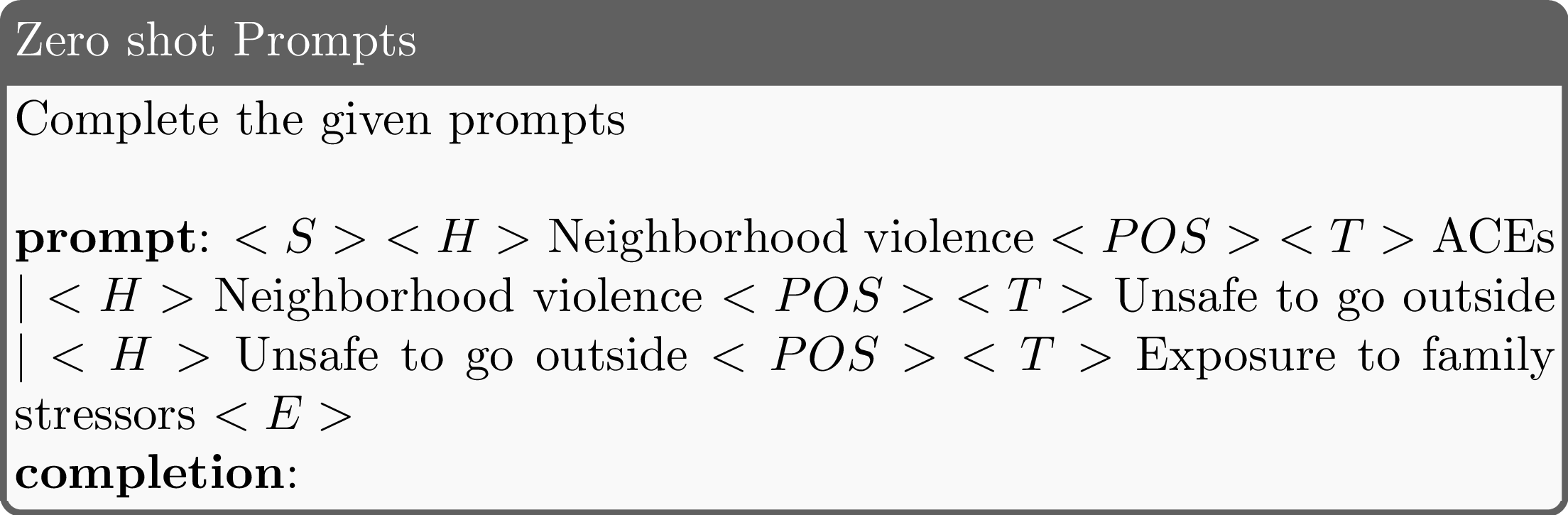}
    \caption{Example of zero-shot prompt.}
    \label{fig:zero}
\end{figure}

\begin{table*}[tbh]
\begin{tabular}{c|p{7cm}|p{0.4\linewidth}}
\toprule
\textbf{Dataset} & \textbf{Input} & \textbf{Text Description} \\ 
\midrule
\multirow{2}{*}{Suicide} & \texttt{$\langle $S$ \rangle$$\langle$H$\rangle$ ACEs of parents $\langle$POS$\rangle$ $\langle$T$\rangle$ Parental risk factors$ \langle$E$\rangle$} & More ACEs of parents increases parental risk factors.\\
 & \texttt{$\langle$S$\rangle$$\langle$H$\rangle$Peers you can talk to$\langle$POS$\rangle$$\langle$T$\rangle$Protective environment$\langle$E$\rangle$} & Having more peers you can talk to can create protective environment.\\
\midrule

\multirow{2}{*}{Obesity} & \texttt{$\langle $S$ \rangle$$\langle$H$\rangle$ Obesity awareness programs$\langle$POS$\rangle$ $\langle$T$\rangle $nutrition education$\langle$H$\rangle $Obesity awareness programs$ \langle$POS$\rangle $community partnerships$ \langle$E$\rangle$} & Obesity awareness programs can develop knowledge about nutrition and also community partnerships.\\
& \texttt{$\langle $S$ \rangle$$\langle$H$\rangle$ routine practices in hospital $\langle$NEG$\rangle$ $\langle$T$\rangle $breastfeeding knowledge$ \langle$E$\rangle$} & Improving routine practices in hospitals decreases breastfeeding knowledge. \\
\bottomrule
\end{tabular}
\caption{Sample instances from the causal graph datasets}
\label{tab:sample_instances}
\end{table*}

The third setting is \textbf{zero-shot}, where the model is only given a natural description of the task along with a test query (see Figure~\ref{fig:zero}). This is the most challenging setting, as it tests whether PLMs have encoded causal relationships between entities without needing any kind of domain-specific support from the user. 

GPT models have one key parameter, known as \textbf{temperature}. Intuitively, it controls the `creative randomness' of the model. When temperature is low, outputs will be less varied because the model will always output the words that have the highest probability. As temperature is increased, the model can select words that have a lower probability, thus leading both to more varied outputs but also to an increased risk of being offtopic. For each of the $3 \times 2$ configurations and four GPT models, we performed experiments on two different temperature levels (0.6 and 0.8). As described in the next section, each experiment took place on two different causal datasets, to measure the effect of the application domain onto the results. In summary, we have a total of $2 \text{ inputs} \times 3 \text{ learning settings} \times 2 \text{ temperatures} \times 2 \text{ datasets}$, that is, 24 experiments.

\subsection{Datasets}

We used two causal maps provided on open repositories: a map on {\em youth suicide} in the U.S. with 361 nodes and 946 edges (\cite{giabbanelli2022pathways}) and a smaller map on {\em obesity} with 98 nodes and 177 edges (\cite{drasic2015exploring}). The maps are available at \url{https://osf.io/7nxp4/} and \url{https://osf.io/7ztwu/}, respectively. The youth suicide map was created by interviewing 15 experts, while the obesity map was developed by 19 experts. In both cases, experts representing a diverse array of fields were interviewed one-on-one, and their individual maps were merged to arrive at the final map. The merging process ensures that a concept appears only once in the entire map. We divided each map into small parts of 2 up to 4 nodes so that each part can be described in a sentence of tolerable length. To create a training dataset for each map, we employed a team of 8 human annotators who independently wrote descriptive sentences for each part. Sentences were then extracted and the list was reduced to promote variations in style. As a result, the generated dataset includes 588 graph-sentence pairs for the suicide map and 625 pairs for the obesity map. Table~\ref{tab:sample_instances} presents sample instances from the datasets, while the dataset splits used in our experiments are shown in Table \ref{tab:data-split}.

\begin{table}[t]
\centering
\small
\begin{tabular}{cccc}
\toprule
\textbf{Dataset}  & Train  & Validation  & Test  \\ \midrule
{Suicide}  &  328 & 83 & 177 \\
{Obesity}  & 349 & 88 & 188 \\    
\bottomrule
\end{tabular}
\caption{Statistics of the datasets.}
\label{tab:data-split}
\end{table}


\begin{table*}[tbh]\centering
\small
\renewcommand{\arraystretch}{1}
\setlength{\tabcolsep}{8.5pt}
\begin{tabular}{cll|cc|cc|cc|cc}\toprule
\multirow{2}{*}{\textbf{Dataset}} &\multirow{2}{*}{\textbf{Training}} &\multirow{2}{*}{\textbf{Model}} &\multicolumn{2}{c}{\textbf{RougeL}} &\multicolumn{2}{c}{\textbf{METEOR}} &\multicolumn{2}{c}{\textbf{BERTScore}} &\multicolumn{2}{c}{\textbf{QuestEval}}  \\
\cmidrule{4-11}

& & &\textbf{Tags} &\textbf{NoTags} &\textbf{Tags} &\textbf{NoTags} &\textbf{Tags} &\textbf{NoTags} &\textbf{Tags} &\textbf{NoTags}\\
\midrule

\multirow{10}{*}{\textbf{Obesity}} &\multirow{3}{*}{\textbf{FT}} & Ada &0.583&	0.577&	0.426	&0.413	&0.977	&0.976	&0.588	&0.561\\
&&Babbage & 0.588	&0.585	&0.433	&0.427	&0.977&	0.976&	0.561	&0.562\\
&& Curie & 0.588	&0.580	&0.440	&0.423	&0.977	&0.976	&0.594	&0.574\\
&& Davinci & \textbf{0.665}	&0.601	&0.444	&{0.444}	&\textbf{0.978}	&0.977	&\textbf{0.602}	&0.582\\

\cmidrule{2-11}
&\multirow{3}{*}{\textbf{Few}} & Ada & 0.298&	0.312	&0.231	&0.279	&0.967&	0.966&	0.416	&0.391\\
&&Babbage & 0.339	&0.311&	0.306	&0.297	&0.967	&0.965&	0.367&	0.406\\
&& Curie& 0.486	&0.434	&0.381	&0.359	&0.973	&0.971	&0.465 &0.476\\
&& Davinci & \textbf{0.575}	&0.541&	\textbf{0.455}&	0.432&	\textbf{0.975}&	0.974&	\textbf{0.602}	&0.582\\
						
	\cmidrule{2-11}					
&\multirow{3}{*}{\textbf{Zero}} & Ada &	0.201&	0.199	&0.188	&0.184	&0.957	&0.957&	0.298	&0.284\\
&& Babbage & 0.338&	0.328&	0.266	&0.267	&0.964	&0.964&	0.353	&0.360\\
&& Curie & 0.401	&0.390	&0.279	&0.278	&0.964	&0.963&	0.359	&0.349\\
&& Davinci & 0.422&	\textbf{0.426}&	0.361	&\textbf{0.385}	&0.967	&\textbf{0.969}&	0.413	&\textbf{0.427}\\

\midrule

\multirow{10}{*}{\textbf{Suicide}} &\multirow{3}{*}{\textbf{FT}} & Ada &0.520	&0.551	&0.241&	0.284	&0.959	&0.961	&0.637	&0.606\\
&& Babbage &0.671	&0.665&	0.489&	0.488	&0.981&	0.981	&0.646	&0.601\\
&& Curie & 0.665&	0.669	&0.483	&0.494	&0.981	&0.981	&0.647	&0.624\\
&& Davinci &\textbf{0.681}&	0.677	&\textbf{0.499}	&0.490	&\textbf{0.982}	&0.981&	\textbf{0.649}	&0.627\\

\cmidrule{2-11}
&\multirow{3}{*}{\textbf{Few}} & Ada & 0.270&	0.226&	0.205&	0.184	&0.967	&0.956	&0.353	&0.337\\
&& Babbage &0.491	&0.464&	0.347&	0.348&	0.974&	0.974&	0.536&	0.417\\
&& Curie & 0.547&	0.612&	0.422	&0.432	&0.976&	0.978	&0.569	&0.578\\
&& Davinci &\textbf{0.617}	&0.529	&\textbf{0.473}&	0.434	&\textbf{0.977}&	0.973&	\textbf{0.649}	&0.627\\
						
	\cmidrule{2-11}					
&\multirow{3}{*}{\textbf{Zero}} & Ada &	0.186&	0.174	&0.182&	0.177&	0.958	&0.957&	0.284	&0.285\\
&& Babbage &0.383	&0.355	&0.282&	0.268&	0.967&	0.965	&0.369&	0.344\\
&& Curie & 0.463&	0.494	&0.330	&0.344&	0.966	&0.967&	0.359	&0.358\\
&& Davinci &0.370	&\textbf{0.429}	&0.333	&\textbf{0.367}	&0.965	&\textbf{0.967}	&0.358	&\textbf{0.383}\\

\bottomrule
\end{tabular}
\caption{Results for generated sentences when the input is formatted with and without tags, across different training settings (fine-tuned, few-shot, and zero-shot) for four GPT-3 models sorted from smallest (Ada, 350 million parameters) to largest (Davinci, 175 billion parameters).}
\label{tab:main_results}
\end{table*}

\subsection{Evaluation Metrics}
We use several automatic metrics as well as human readers to evaluate the quality of the generated text descriptions. For the \textbf{automatic evaluation}, we relied on widely-used reference-based metrics where the model generated output is compared to the human-written reference text. We considered four evaluation metrics. \textit{ROUGE-L} calculates the longest common subsequence overlap between the human text and model-generated text\footnote{\url{https://github.com/google-research/google-research/tree/master/rouge}}. \textit{METEOR} (Metric for Evaluation of Translation with Explicit ORdering) is also an $n$-gram matching metric, but it accounts for semantics\footnote{\url{https://github.com/wbwseeker/meteor}}. \textit{BERTScore} (\cite{zhang2019bertscore}) focuses on semantic similarity between the reference and candidate texts\footnote{\url{https://github.com/Tiiiger/bert_score}}. Finally, \textit{QuestEval} (\cite{scialom2020QuestEval}) focuses assessing factuality \footnote{\url{https://github.com/ThomasScialom/QuestEval}}, which is an important property for all NLG and especially so for graph-to-text generation; it has a good correlation with human ratings (\cite{li2022faithfulness}). All the evaluation metrics produce scores from 0 to 1, where 1 is the best match between the generated text and the human-written reference text.


We also conduct \textbf{human evaluation} of the outputs to assess two dimensions of quality. First, \textit{faithfulness} measures how much of the input subgraph is reflected in the generated sentence; this score is negatively impacted when the model hallucinates. Second, \textit{coverage} measures how much of the input is preserved in the output; this score is lower when the model operates a simplification by ignoring parts of the input. We invited two annotators and asked them to choose the best sentence for faithfulness and coverage over 20 samples. This was repeated for the two datasets, the two harder training settings (fine-tuned, zero-shot), the two forms of input (with/without causality), and the best model (i.e., the GPT-3 model with best performance on automatic metrics). To reduce the risk that annotators may miss information when reading an input as linearized text, we also generated causal graphs for each input in the same format as Figure \ref{fig:Obesity_causal}. We utilized Cohen's Kappa score to calculate the inter-annotator agreement score for faithfulness and coverage in both datasets and for each training setting.

\section{Results}
\label{sec:results}
Table~\ref{tab:main_results} shows the main results of our experiment to assess whether formatting the input linearized representation with and without tags improves the model's generated sentences, depending on the dataset, GPT-3 model size, and automatic metric. All results are presented at a temperature setting of 0.6 due to space limitation, while noting that similar results were obtained at a higher temperature of 0.8. Davinci is the best model in all cases. Although results show that providing tags is generally beneficial, these benefits are limited and depend on additional considerations. Tags are most beneficial in fine-tuning and few-shot settings, but not having tags is best in a zero-shot learning situation. This observation is supported across the two datasets and across metrics, at the exception of measuring fine-tuning on obesity with METEOR (performances with and without tags are tied). As expected, results deteriorate as we move from fine-tuning (best results) to few-shot and then zero-shot (worst results). Interestingly, we observe only a minor deterioration when shifting from a full-training dataset to using just three instances, and a more pronounced decline when moving to a zero-shot setting. 

We complemented this automatic analysis by performing a human evaluation for the results obtained by the best model, Davinci. The inter-annotator agreement score (Table~\ref{tab:interannotator}) shows moderate to very strong agreement under a zero-shot setting, but poor agreement between dimensions (faithfulness and coverage) in both datasets for fine-tuning. These varying levels of agreement are routinely observed in the literature (\cite{ethayarajh2022human}), as identifying robust mechanisms for manual evaluation remains an active area of research. Within these limitations, results (Table~\ref{tab:freq_tags_notags}) show that human evaluators prefer the text generated by models using tags in all but one case (zero-shot learning for obesity). This manual inspection confirms the takeaway of the automatic metrics: guiding GPT-3 with causal tags leads to higher performance in general, but not necessarily under a zero-shot setting. 


\begin{table}[t]
\centering
\small
\begin{tabular}{ccc|cc} 
 \toprule
  & \multicolumn{2}{c}{Faithfulness} & \multicolumn{2}{c}{Coverage}  \\
  \cmidrule{2-5}
  & FT & ZS & FT & ZS\\
 \midrule
 Suicide & -0.19 & 0.33 & -0.19 & 1\\
 Obesity & -0.5 & 0.57 & 0 & 0.39 \\
 \bottomrule
\end{tabular}
\caption{Inter-annotator (Cohen's Kappa) agreement score for faithfulness and coverage}
\label{tab:interannotator}
\end{table}

\begin{table}[t]
\centering
\small
\begin{tabular}{ccc|cc} 
 \toprule
  & \multicolumn{2}{c}{Faithfulness} & \multicolumn{2}{c}{Coverage}  \\
  \cmidrule{2-5}
  & Tags & NoTags & Tags & NoTags\\
 \midrule
 Obesity FT & 69.23 & 30.77 & 65.38 & 34.62\\
 Obesity ZS & 69.23 & 30.77 & 46.15 & 53.85\\
 \midrule
  Suicide FT & 57.69 & 42.32 & 61.54 & 38.46 \\
 Suicide ZS & 69.23 & 30.77 & 73.08 & 26.92\\
 \bottomrule
\end{tabular}
\caption{Results (percentage) of human annotation of comparing Tags vs. NoTags models}
\label{tab:freq_tags_notags}
\end{table}

\begin{table*}[tbh]\centering
\small
\renewcommand{\arraystretch}{1.2}
\setlength{\tabcolsep}{8.5pt}
\begin{tabular}{llcc|cc|cc|cc}\toprule

{\textbf{Training}} &{\textbf{Model}} &\multicolumn{2}{c}{\textbf{RougeL}} &\multicolumn{2}{c}{\textbf{METEOR}} &\multicolumn{2}{c}{\textbf{BERTScore}} &\multicolumn{2}{c}{\textbf{QuestEval}} \\
\cmidrule{2-10}

& & \textbf{Tags} &\textbf{NoTags} &\textbf{Tags} &\textbf{NoTags} &\textbf{Tags} &\textbf{NoTags} &\textbf{Tags} &\textbf{NoTags}\\
\midrule

\multirow{3}{*}{FT}	& WebNLG & 0.396 &	0.316 &	0.321	&0.278	&0.977&	0.980 & 0.514 &	0.573\\
& Obesity	&0.665	&0.601	&0.444	&0.444	&0.978	&0.977 & 0.602 & 0.582\\
& Suicide	&\textbf{0.680}	&{0.676}	&\textbf{0.499}	&{0.490}	&\textbf{0.981}	&{0.981} & \textbf{0.649} & 0.627\\
\midrule
						
\multirow{3}{*}{FS} & WebNLG &	0.399	&0.374	&0.185	&0.282	&0.975	&{0.975} & 0.510 &	0.456\\
& Obesity	&0.575	&{0.541}	&0.455	&0.432	&0.975	&0.974 & 0.602 & 0.582\\
& Suicide	&\textbf{0.617}	&0.528	&\textbf{0.472}	&{0.434}	&\textbf{0.977}	&0.973 & \textbf{0.649} & 0.627\\
\midrule
						
\multirow{3}{*}{ZS} & WebNLG &	\textbf{0.455}	&0.361	&0.316	&0.197	&\textbf{0.980}	&{0.973} & \textbf{0.573} &	0.421\\
& Obesity&	0.422	&0.426&	{0.361}&	\textbf{0.385}	&0.967	&0.969 & 0.413 & 0.427\\
& Suicide&	0.369	&{0.428}	&0.332&	0.366	&0.964	&0.966 & 0.358 & 0.383\\
\bottomrule
\end{tabular}
\caption{Comparing WebNLG against two causal datasets (Obesity, and Suicide) formatted with/without tags, across different training settings (full-shot, few-shot, and zero-shot) for GPT-3 (Davinci, with temperature 0.6).}
\label{tab:webnlg}
\end{table*}

To further examine these results, we compared the performances of our best model (Davinci at temperature 0.6) in our two causal datasets of suicide and obesity against performances in the classical WeBNLG dataset for graph-to-text task, which encodes facts between entities rather than the type of causality. The results (Table~\ref{tab:webnlg}) indicate that causal datasets lead to better performance than WebNLG when fine-tuning or in a few-shot learning setting but, interestingly, WebNLG outperforms in a zero-shot setting. This suggests that while causal relationships may be relatively easier for models to learn with limited training data, they do not appear to be {\em encoded} inherently in the PLM.

\section{Discussion and Conclusions}
\label{sec:discussion}

This paper evaluates text generation specifically for causal graph representations. Using several versions of GPT-3 models and two causal datasets (obesity and suicide), we assess the models' ability to generate natural language descriptions with and without causal tags. The generated outputs are evaluated through both automatic and manual evaluation. Our results indicate that the quality of text generated from a causal map is about the same when using a full training set compared to just three examples. This is an important finding, as creating extensive training sets is particularly labor-intensive, and users would thus be able to save significant amounts of time in exchange for a small loss in performance. Zero-shot learning is a very different setting, which shows a sharp deterioration in performance and an interesting reversal since models learned best \textit{without} using causal tags. Comparing results between causal datasets and the WebNLG dataset suggest that the generative AI tool GPT-3 is able to satisfactorily learn causality with limited training data, but it does not inherently encode causality. 


There are three main limitations to this study. First, creating a full training set for a given causal map is a labor intensive process, hence only two datasets were available for the evaluation. As other research groups gradually examine the use of LLMs for causal maps and share their datasets, additional evaluations will become possible and will contribute to assessing generalizability. Second, although our results were in agreement between the two datasets for automatic metrics, we noted one discrepancy when involving human annotators -- a process that is itself subject to considerable variability. Third, our results focused on evaluating the quality of \textit{sentences}, but reports consist of paragraphs. Extending sentence-level scores over paragraphs could be realized by using the average score of individual sentences in a paragraph, but that would not evaluate the flow. Paragraph-level metrics such as Flesch-Kincaid scores have been employed for NLG, but additional metrics are needed to capture cohesiveness and factuality at the paragraph level. There is also a need for causality-specific metrics that go beyond overlap, semantic similarity, or n-gram matching. Existing metrics can score highly for texts expressing opposing causal relationships, hence we need a more precise assessment of the causal direction, type, and relationships between entities.

Our study focused on causal reasoning for \textit{general} facts, such as the notion that an increase in traumatic events does raise the average risk of suicidal ideation across individuals. Our study of causal reasoning could thus be extended to gain a better understanding of the specific context of a user. For instance, \textit{emotional causality} allows to relate the feelings expressed by a user to their underlying causes, which results in a more empathetic interaction. This also involves a knowledge graph, automatic evaluations (e.g., BLEU) and manual evaluations (e.g., fluency). The main differences would be about the content of the graph and the incorporation of an addition manual evaluation on the empathy expressed in the generated content (\cite{wang2021empathetic}).

Finally, we examined whether LLMs could determine the causal type of a specified relation. Mathematically, we provide the structure (including the direction of each edge) and check for the polarity of the edges' labels. The ability of a LLM to perform this task is promising to support text generation focused on intervention (e.g., does taking an aspirin increase or decrease my headache?). However, this is a simpler task than determining the \textit{direction} of each relation (e.g., $A \rightarrow B$ vs. $B \rightarrow A$), the \textit{level} of causality (e.g., necessary, sufficient), or even discovering the full graph (\cite{kiciman2023causal}). Additional research on such advanced tasks is necessary to support retrospective reasoning (e.g., why is my headache gone?), which is at the forefront of debates on the capacity of generative AI (\cite{bishop2021artificial}).

\small
\printbibliography 

\normalsize

\section*{Appendix: Pseudocode} \label{sec:appendix}

\begin{algorithm}[!b]\small
\caption {Make N-shot Samples} \label{alg:n-shot}
\SetKwInput{KwData}{Variables}
\SetKwFunction{FS}{MakeNShotSamples}
\KwInput{$trainPath$: Path to training data, \\
\hspace{1.1cm} $n$: Number of input instances to select from the dataset}
\KwOutput{$prompt$: Generated prompt based on $n$}
\KwData{$trainFrame$: Dataframe containing training data, $sampled \textunderscore data$: Dataframe containing $n$ randomly sampled instances from $trainFrame$}
\SetKwProg{Fn}{Function}{}{}
  \Fn{\FS{trainPath, $n$}}{
  \tcc{Read dataframe from the path}
  $trainFrame \gets \text{pd.read\textunderscore csv(trainPath)}$ \\
  \tcc{Randomly sample $n$ instances from the dataframe}
  $sampled \textunderscore data \gets trainFrame.sample(n)$ \\
  \tcc{OpenAI GPT3 requires initial statement to give some information about the task.}
  $statement  \gets \textit{``Complete the given prompts''} + \textit{``\textbackslash n\textbackslash n''}$ \\
  \tcc{Initialize prompt as empty string, populated later}
  $\text{prompt} \gets \textit{``''}$ \\
  \tcc{Initializing separator as required by OpenAI GPT3}
  $\text{separator} \gets \textit{``\textbackslash n \textbackslash n \#\#\# \textbackslash n\textbackslash n''}$ \\
  \tcc{Iterate over sampled dataframe and create the prompt.}
  \For{row in sampled\textunderscore data.iterrows()}{
    $(sentence, completion) \gets \textit{row[``prompt''], row[``completion'']}$ \\
    \tcc{Replace $\langle$end$\rangle$ tokens.}
    $completion \gets \textit{completion.replace(``$\langle$end$\rangle$'', ``'')}$
    $prompt$ = $prompt$ + \textit{(
                        ``prompt: '' \\
                        + $sentence$ + ``\textbackslash n'' + ``completion: '' \\
                        + completion + separator )}\\
                    
    }
    \Return $statement + prompt$
}
\end{algorithm} 

\begin{algorithm}[!ht]\small
\caption {Generate response of openAI GPT3 models} \label{alg:get-response}
\SetKwInput{KwData}{Variables}
\SetKwFunction{GS}{getResponse}
\KwInput{$testPath$: path to testing data, $temperature$: setting used to generate the outputs, $model$: name of openAI model to use, $maxTokens$: number of new tokens to generate, $trainPath$: path to the training dataset, $n$: number of input instances to select from the dataset}
\KwOutput{$results$: A dictionary consisting of outputs generated by specified OpenAI GPT3 model.}
\SetKwProg{Fn}{Function}{}{}
  \Fn{\GS{$trainPath$, $n$, $testPath$, $temperature$, $model$, $maxTokens$}}{
  \tcc{Get few shot input prompt}\
  $prompt \gets GenerateNShotSamples(trainPath, n)$\\
  \tcc{Read dataframe from the path} \
  $testFrame \gets pd.read\textunderscore csv(testPath)$ \\
  \tcc{Initialize empty dictionary to store the results.}
  $results \gets \phi$ \\
  \tcc{Iterate over the test prompts.} \
  \For{testPrompt in $testFrame[\textit{``prompt''}]$}{
   $inputPrompt$ = $prompt$ + \textit{``prompt: ''} \\ 
   + $testPrompt$ + \textit{``\textbackslash n''} +\textit{``completion: ''}  \\
   + \textit{``\textbackslash n \textbackslash n''} \\
    \tcc{Generate the output using OpenAI API.}
   $response \gets openai.Completion.create(model=$\\
	$model$,$prompt = inputPrompt$, \\
   
   \textit{$max$\textunderscore $tokens$} = $maxTokens$, \\
   $temperature$ = $temperature$) \\
   $results \gets response$
  }
  \Return $results$
}
\end{algorithm} 

The process of generating n-shot prompts is described in Algorithm \ref{alg:n-shot}. The $MakeNShotSamples$ function takes two arguments as inputs, namely the path to training data ($trainPath$) and  the number of input instances to select from the training data ($n$), i.e., the number of n-shot samples. The function $MakeNShotSamples$ generates the n-shot prompt in the format required by the OpenAI API. Lines 2 and 3 show how to read the dataframe from the given path and randomly sample $n$ instances from the dataframe. OpenAI GPT3 models require a statement that indicates what task needs to be performed and is included in line 4. Finally, lines 7-16 populate the $prompt$ variable (line 5) by iterating over the randomly sampled $n$ instances and concatenating them in the format required by the API.

The function $getResponse$ (Algorithm \ref{alg:get-response}) accepts multiple parameters and queries OpenAI API to get the response for each prompt in the test set. We initialize an empty dictionary ($results$) which stores the outputs generated by the model. Lines 5-16 show the process of iterating over test samples and querying the OpenAI API with the required parameters. Finally, after the results are generated for all the samples in the test set, the function returns all results in a dictionary.



\end{document}